\documentclass[10pt,twocolumn,letterpaper]{article}

\usepackage{cvpr}
\usepackage{times}
\usepackage{epsfig}
\usepackage{graphicx}
\usepackage{amsmath}
\usepackage{amssymb}

\usepackage{color}
\usepackage{enumerate}

\usepackage[breaklinks=true,bookmarks=false]{hyperref}

\cvprfinalcopy 

\def\cvprPaperID{1523} 

\ifcvprfinal\fi
\begin{document}

\newcommand{\vitto}[1]{\textcolor{red}{VF: [#1]}}
\newcommand{\dimpp}[1]{\textcolor{magenta}{DP: [#1]}}
\definecolor{myMagenta}{rgb}{1, 0, 1}
\newcommand{\updated}[1]{\textcolor{myMagenta}{#1}}
\newcommand{\added}[1]{\textcolor{myMagenta}{added: [#1]}}

\title{How hard can it be? Estimating the difficulty of visual search in an image}


\author{Radu Tudor Ionescu$^1$, Bogdan Alexe$^{1,4}$, Marius Leordeanu$^3$, Marius Popescu$^1$,\\
Dim P. Papadopoulos$^2$, Vittorio Ferrari$^2$\\
\\
\vspace{-0.2cm}
{$^1$University of Bucharest,\;\;\;\;\;\;\;\;\;$^2$University of Edinburgh,}\\
\and
{$^3$Institute of Mathematics of the Romanian Academy,}\\
\and
{$^4$Institute of Mathematical Statistics and Applied Mathematics of the Romanian Academy}\\
}

\maketitle

\begin{abstract}
\vspace{-0.2cm}
We address the problem of estimating image difficulty defined as the human response time for solving a visual search task. We collect human annotations of image difficulty for the PASCAL VOC 2012 data set through a crowd-sourcing platform. We then analyze what human interpretable image properties can have an impact on visual search difficulty, and how accurate are those properties for predicting difficulty. Next, we build a regression model based on deep features learned with state of the art convolutional neural networks and show better results for predicting the ground-truth visual search difficulty scores produced by human annotators. Our model is able to correctly rank about $75\%$ image pairs according to their difficulty score. We also show that our difficulty predictor generalizes well to new classes not seen during training. Finally, we demonstrate that our predicted difficulty scores are useful for weakly supervised object localization ($8\%$ improvement) and semi-supervised object classification ($1\%$ improvement).
\end{abstract}

\vspace{-0.5cm}
\section{Introduction}
\vspace{-0.1cm}

\begin{figure*}[t]

\begin{center}
\includegraphics[width=1.0\linewidth]{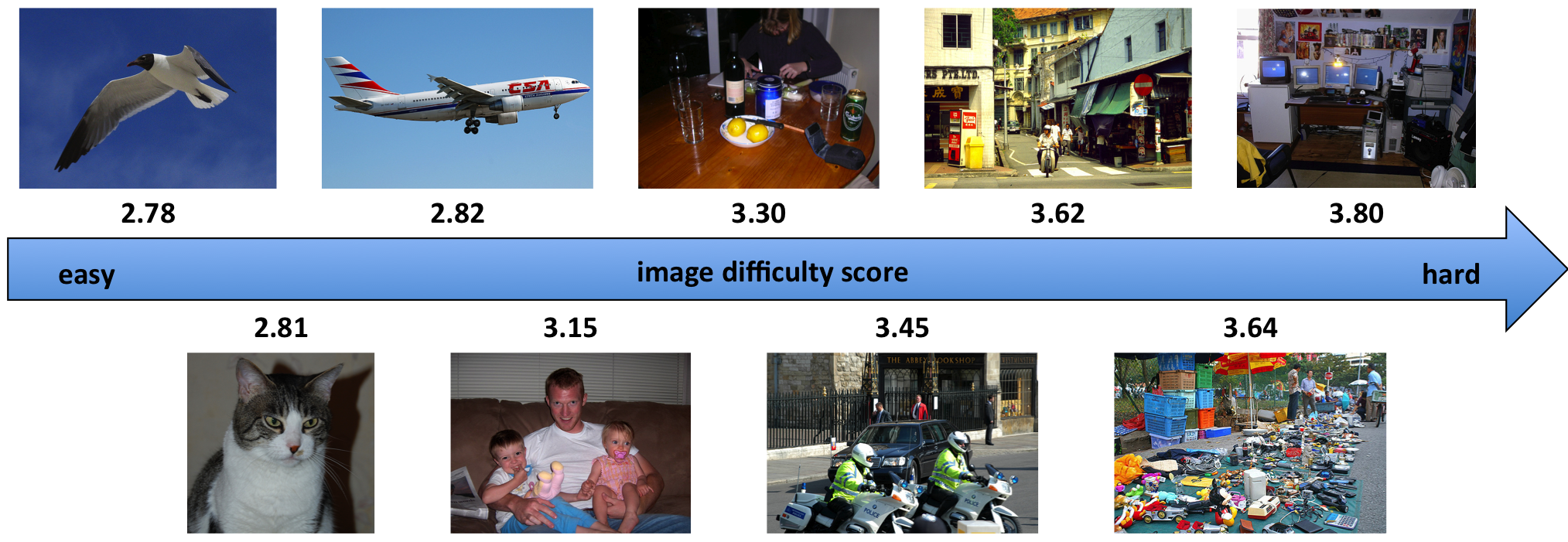}
\end{center}
\vspace*{-0.3cm}
\caption{Images with difficulty scores predicted by our system in increasing order of their difficulty.}
\label{fig_easy_vs_medium_vs_hard}
\vspace*{-0.3cm}
\end{figure*}

Humans can naturally understand the content of images quite easily. The visual human perception system works by first recognizing the 'gist' of the image almost instantaneously~\cite{Oliva2005,Oliva2001}, just from a single glance (200 ms) and, then, in a second stage, by recognizing the individual objects in the image~\cite{Oliva2001} as a result of visual search. Cognitive studies~\cite{Arun-2012,Trick-1998,Wolfe-2010} show evidence that, for the task of searching for a pattern in an image, the user response time is proportional to the visual search difficulty, which could vary from one image to another. Images are not equal in their difficulty: some images are easy to search and objects are found fast while others are harder, requiring intensive visual processing by humans. The measure of visual search difficulty could be related to several factors such as background clutter, complexity of the scene, number of objects, whether they are partially occluded or not, and so on. 

In this paper, we address the problem of estimating visual search difficulty. This topic is little explored in the computer vision literature with no data sets assessing the difficulty of an image being available. We approach our study by collecting annotations on the PASCAL VOC 2012 data set~\cite{PASCAL-voc-2012} as human response times during a visual search task and convert them into difficulty scores (Section~\ref{sec_Main}).
While measuring visual search difficulty by human observations might be subject to some user variability, we believe that there are intrinsic image properties that constitute the ingredients in the unknown underlying recipe of making an image difficult (Figure~\ref{fig_easy_vs_medium_vs_hard}). We use the PASCAL VOC 2012 images annotated with difficulty scores to investigate in depth how different image properties correlate with the ground-truth difficulty scores. 
%
We find that higher level features, such as the ones learned with convolutional neural networks (CNN)~\cite{Hinton-NIPS-2012} are the most effective, suggesting that visual search difficulty is indeed a measure that relates to higher level cognitive processing. Using such features, we train models to automatically predict the human assessment of visual search difficulty in an image (Section~\ref{sec_prediction}).
We release the human difficulty scores we collected on PASCAL VOC 2012, as well as our code to predict the difficulty of any image at {http://image-difficulty.herokuapp.com}.

Measuring image difficulty could have many potential applications that use the primary information that some images are harder to analyze than others. In Section~\ref{sec_applications}, we demonstrate the usefulness of our difficulty measure in two object recognition applications. For the task of weakly supervised object localization, we show how to enhance standard methods based on multiple instance learning~\cite{bilen14bmvc,cinbis14cvpr,deselaers-IJCV-2012,shi12bmvc,siva11iccv,song14icml,song14nips} with our measure and obtain an $8\%$ improvement.
Similarly, for the task of semi-supervised object classification, we use our measure to improve the accuracy of a classifier based on CNN features~\cite{Simonyan-ICLR-14}
by $1\%$.

\noindent
{\bf Related work.} There are many computer vision works analyzing global image properties such as saliency~\cite{gao:iccv07,Hou2007CVPR,Itti1998PAMI,marchesotti:iccv09,moosmann:eccv06}, memorability~\cite{Isola2014PAMI, Isola2011CVPR}, photo quality~\cite{Luo2008ECCV} and objects' importance~\cite{Spain2008ECCV}.
However, there is little work on the topic of image difficulty~\cite{Dingding2011MLDM,Russakovsky2015,Vijayanarasimhan2009CVPR}.  
%
Russakovsky et al.~\cite{Russakovsky2015} measure difficulty as the rank of an object's bounding-box in the order of image windows induced by the objectness measure~\cite{objectness-CVPR-2010,objectness-PAMI-2012}. This basically measures image clutter. However, it needs ground-truth bounding-boxes in order to quantify difficulty (even at test time).
%
Liu et al.~\cite{Dingding2011MLDM} predict the performance of a segmentation algorithm to be applied to an image, based on various features including gray tone, color, gradient and texture (on just $100$ images).
%
More closely related to our idea, Vijayanarasimhan and Grauman~\cite{Vijayanarasimhan2009CVPR} try to predict the difficulty of an image in terms of the time needed by a human to segment it, with the specific goal of reducing manual annotation effort. They select candidate low-level features and train multiple kernel learning models to predict easy versus hard images.
However, the image segmentation task~\cite{Vijayanarasimhan2009CVPR} is conceptually different from our visual search task. For example, it might be very easy to find a tree in a particular image, although it can be very hard to segment, while a truncated car can be easily segmented but difficult to find and recognize.
Jain and Grauman~\cite{Dutt_2013_ICCV} predict what level of human annotation will be sufficient for interactive segmentation to succeed. Their approach learns the image properties that indicate how successful a given form of user input will be.

In contrast to these previous works, we approach the problem from a higher level of image interpretation, for the general task of visual search and collect annotations for a much larger data set of over 10K images. 

\section{Image difficulty from a human perspective}
\label{sec_Main}
\vspace{-0.1cm}

Supported by cognitive studies~\cite{Arun-2012,Trick-1998,Wolfe-2010}, we consider that the difficulty of an image is related to how hard it is for a human to decide the presence or absence of a given object class in an image.
We quantify the difficulty as the time needed by a human to solve this visual search task. This value could depend on several factors such as the amount of irrelevant clutter in the image, the number of objects, their scale and position, their class type, the relevant contextual relationships among them, occlusions and other kinds of noise. We thoroughly investigate how these properties correlate with the visual search difficulty in Section~\ref{sec_what_makes}.

First, we designed a visual search protocol for collecting human response times on a crowd-sourcing platform, namely CrowdFlower\footnote{http://www.crowdflower.com/}. We collected ground-truth difficulty annotations by human evaluators on a per image basis for all $11,540$ \emph {train} and \emph{validation} images in PASCAL VOC 2012 data set~\cite{PASCAL-voc-2012}. 
This data set contains images with object instances from $20$ classes (\emph{aeroplane}, \emph{boat}, \emph{cat}, \emph{dog}, \emph{person} and so on) annotated with bounding-boxes.
The images vary in their difficulty: objects appear against a variety of backgrounds, ranging from uniform to heavily cluttered, and vary greatly in their number, location, size, appearance, viewpoint and illumination. This variety makes this data set very suitable for collecting ground-truth difficulty annotations.
We next describe the protocol and present informative statistics about the collected data.

\subsection{Can we measure visual search difficulty?}
\label{ssec_measuring}
\vspace{-0.1cm}

\noindent
{\bf Collecting response times.}
We collected ground-truth difficulty annotations by human evaluators using the following protocol: (i) we ask each annotator a question of the type ``Is there an \emph{\{object class\}} in the next image?'', where \emph{\{object class\}} is one of the $20$ classes included in the PASCAL VOC 2012; (ii) we show the image to the annotator; (iii) we record the time spent by the annotator to answer the question by ``Yes'' or ``No''. Finally, we use this response time to estimate the visual search difficulty.

To make sure the measured time is representative, the annotator has to signal that he or she is ready to see the image by clicking a button (after reading the question first). After seeing the image and analyzing it, the annotator has to signal when he or she made up his mind on the answer by clicking another button. At this moment we hide the image to prevent cheating on the time. Moreover, we made sure the annotation task is not trivial by associating two questions for each image, such that the ground-truth answer for one question is positive (the object class specified in the question \emph{is present} in the image) and the ground-truth answer for the other question is negative (the object class specified in the question \emph{is not present} in the image). In this way we prevented a bias in obtaining answers uncorrelated with the image content, constraining the annotator to be focused during the entire task. Each answer (``Yes'' or ``No'') has a $50\%$ chance of being the right choice. Naturally, an annotator could memorize an image and answer more quickly if the image would be presented several times, so we made sure that a person did not get to annotate the same image twice. Each question was answered by three human annotators. Given that we used $11,540$ images and we associated two questions per image, we obtained $69,240$ annotations. The annotations come from $736$ trusted contributors. A \emph{trusted} contributor has an accuracy (percentage of answers that match the ground-truth answers) higher than $90\%$.


\noindent
{\bf Data post-processing and cleanup.}
When the annotation task was finished, we had $6$ annotations per image ($3$ for each of the two questions) with the associated response times. We removed all the response times longer than $20$ seconds, and then, we normalized each annotator's response times by subtracting the annotator's mean time and by dividing the resulted times by the standard deviation.  
We removed all the annotators with less than $3$ annotations since their mean time is not representative. We also excluded all the annotators with less than $10$ annotations with an average response time higher than $10$ seconds. After removing all the outliers, the difficulty score per image is computed as the geometric mean of the remaining times. It is worth mentioning that by adjusting the accuracy threshold for trusted annotators to $90\%$, we allow some wrong annotations in the collected data. 
Wrong annotations provide the ultimate evidence of a difficult image, showing also that the problem of estimating image difficulty is not trivial. We determined the images containing wrong annotations (based on the ground-truth labels from PASCAL VOC 2012) and added a penalty to increase the difficulty scores of these images.

\begin{table}[!t]
\footnotesize{
\begin{center}
\begin{tabular}{|l|c|c|c|}
\hline
                & Mean                  & Minimum   & Maximum\\
\hline
\hline
Kendall $\tau$  & $0.562 \pm 0.127$     & $0.182$    & $0.818$\\
\hline
\end{tabular}
\end{center}
}
\vspace*{-0.1cm}
\caption{Kendall's $\tau$ rank correlation coefficient among $58$ trusted annotators, on a subset of $56$ images. The response time of each annotator is compared to the mean response time of all annotators.}
\label{tab_human_corr}
\vspace*{-0.3cm}
\end{table}

\noindent
{\bf Human agreement.}
We report the inter-human correlations on a subset of $56$ images that we used to spot untrusted annotators in CrowdFlower. We consider only the $58$ trusted annotators who annotated all these $56$ images. In this setting, we compute the correlation following a one-versus-all scheme, comparing the response time of an annotator to the mean response time of all annotators. For this, we use the Kendall's $\tau$ rank correlation coefficient~\cite{Kendall-1948, upton-dict-stat-2008}. Kendall's $\tau$ is a correlation measure for ordinal data based on the difference between the number of concordant pairs and the number of discordant pairs among two variables, divided by the total number of pairs. The mean Kendall's $\tau$ correlation is reported in Table~\ref{tab_human_corr}, along with the standard deviation, the minimum and the maximum correlations obtained. The mean value of $0.562$ means that the average human ranks about $80\%$ image pairs in the same order as given by the mean response time of all annotators.
This high level of agreement among humans demonstrates that visual search difficulty can indeed be consistently measured.

\subsection{What makes an image difficult?}
\label{sec_what_makes}
\vspace{-0.1cm}

Images are not equal in their difficulty. In order to gain an understanding of what makes an image more difficult than another, we consider several human interpretable image properties and analyze their correlation with the visual search difficulty assessed by humans. The image properties are derived from the human manual annotations provided for each image in PASCAL VOC 2012~\cite{PASCAL-voc-2012}. 
All object instances of the $20$ classes are annotated with bounding boxes and other several details (viewpoint, 
truncation, occlusion, difficult flags) regarding the annotated object (for more details see~\cite{PASCAL-voc-2012}).
In our analysis, we consider the following image properties:
(i) number of annotated objects;
(ii) mean area covered by objects normalized by the image size;
(iii) non-centeredness, defined as the mean distance of the center of all objects' bounding boxes to image center normalized by the square root of image area;
(iv) number of different classes;
(v) number of objects marked as truncated;
(vi) number of objects marked as occluded;
(vii) number of objects marked as difficult.

It is important to remark that these image properties are not available at test time. We only use them in our analysis to study how human interpretable properties correlate with visual search difficulty and also how well these properties could predict difficulty.

\begin{table}[t]
\footnotesize{
\begin{center}
\begin{tabular}{|cl|r|}
\hline
        & Image property                        &  Kendall $\tau$ \\
\hline
\hline
(i)     & number of objects                     & $0.32$ \\
(ii)   & mean area covered by objects          & $-0.28$ \\
(iii)    & non-centeredness                      & $0.29$ \\
(iv)     & number of different classes           & $0.33$ \\
(v)    & number of truncated objects           & $0.22$ \\
(vi)   & number of occluded objects            & $0.26$ \\
(vii)  & number of difficult objects           & $0.20$ \\
\hline
(viii)    & combine (i) to (vii) with $\nu$-SVR  & $0.36$ \\
\hline
\end{tabular}
\end{center}
}
\vspace*{-0.1cm}
\caption{Kendall's $\tau$ rank correlations for various image properties.}
\label{tab_image_properties}
\vspace*{-0.4cm}
\end{table}

We quantify the correlation between image properties and visual search difficulty assessed by humans (Section~\ref{ssec_measuring}) by measuring how well image properties scores can predict ground-truth human difficulty scores. More precisely, we compute the Kendall's $\tau$ correlation between the rankings of the images when ranked either by the image properties scores or by the ground-truth human difficulty scores.
Each image property assigns visual search difficulty scores in a range that is different from the range of the ground-truth scores. Kendall's $\tau$ is suitable for our analysis because it is invariant to different ranges of the various measurements.

In all our experiments on visual search difficulty prediction throughout this paper, we divided the $11,540$ samples included in the official training and validation sets of PASCAL VOC 2012 into three subsets. We used $50\%$ of the samples for training, $25\%$ for validation and another $25\%$ for testing. 
Table~\ref{tab_image_properties} shows the Kendall's $\tau$ rank correlations between the difficulty scores based on the image properties and the ground-truth difficulty scores on our test set.
The results confirm that human interpretable properties are informative for predicting visual search difficulty. The top three most correlated image properties with the ground-truth difficulty score specify some of the ingredients that make an image difficult: the image should contain many instances of different classes scattered all over the image (not just in the center). The next most informative property is the mean area covered by objects. It shows a negative correlation with the ground-truth difficulty score suggesting that, on average, small objects are more difficult to find.
Interestingly, difficulty could also be predicted to some degree based on the number of objects marked as truncated, occluded or difficult. However, as most objects appear normally, without being truncated or occluded, these markers are rarely used, which reduces their predictive power.
As each image property captures a different characteristic, combining them appears to be promising. We trained a Support Vector Regression ($\nu$-SVR) model~\cite{taylor-Cristianini-cup-2004} to combine all seven image properties. In our evaluation, we used the $\nu$-SVR implementation provided in~\cite{libsvm}. The combination yields the highest Kendall's $\tau$ correlation ($0.36$).
In Section~\ref{sec_prediction}, we show that we can learn an even better predictor capable of automatically assessing visual search difficulty based on CNN features, without information derived from image properties. 

\begin{table}[t]
\footnotesize{
\begin{center}
\begin{tabular}{|l|r|r||l|r|r|}
\hline
Class           & Score     & mAP
& Class         & Score     & mAP\\
\hline
\hline
bird            & $3.081$   & $92.5\%$  & bicycle       & $3.414$   & $90.4\%$\\
cat             & $3.133$   & $91.9\%$  & boat          & $3.441$   & $89.6\%$\\   
aeroplane       & $3.155$   & $95.3\%$  & car           & $3.463$   & $91.5\%$\\
dog             & $3.208$   & $89.7\%$  & bus           & $3.504$   & $81.9\%$\\  
horse           & $3.244$   & $92.2\%$  & sofa          & $3.542$   & $68.0\%$\\
sheep           & $3.245$   & $82.9\%$  & bottle        & $3.550$   & $54.4\%$\\ 
cow             & $3.282$   & $76.3\%$  & tv monitor    & $3.570$   & $74.4\%$\\
motorbike       & $3.355$   & $86.9\%$  & dining table  & $3.571$   & $74.9\%$\\
train           & $3.360$   & $95.5\%$  & chair         & $3.583$   & $64.1\%$\\
person          & $3.398$   & $95.2\%$  & potted plant  & $3.641$   & $60.7\%$\\
\hline
\end{tabular}
\end{center}
}
\vspace*{-0.1cm}
\caption{Average difficulty scores per class produced by humans versus the classification mean Average Precision (mAP) performance of the best model presented in~\cite{Chatfield-BMVC-14} for the $20$ classes available in PASCAL VOC 2012. Classes are sorted by human scores.}
\label{tab_scores_per_class}
\vspace*{-0.1cm}
\end{table}

\subsection{Visual search difficulty at the class level}
\vspace{-0.1cm}

We can produce some interesting statistics based on our collection of difficulty scores. Perhaps one of the most interesting aspects is to study the difficulty scores at the class level. We compute a difficulty score per object class by averaging the score for the images that contain at least one instance of that class. The difficulty scores for all the $20$ classes in PASCAL VOC 2012 are presented in Table~\ref{tab_scores_per_class}. It appears that \emph{bird}, \emph{cat} and \emph{aeroplane} are the easiest object classes in PASCAL that can be found in images by humans. We believe that birds and aeroplanes are easy to find as they usually appear in a simple, uniform background, for example on the sky. On the other hand, cats can appear in various contexts (simple or complex), but their distinctive shape, eyes and other body features are probably very easy to recognize. The most difficult classes in PASCAL, from a human perspective, appear to be \emph{potted plant}, \emph{chair}, \emph{dining table} and \emph{tv monitor}. We believe that potted plants and chairs are hard to find due to high (intra-class) variability in their appearance. For instance, chairs come in different shapes and sizes, such as stools, armchairs, and so on. Furthermore, all the difficult classes usually appear in complex contexts, such indoor scenes with many objects and varying illumination conditions. Interestingly, the difficulty scores presented in Table~\ref{tab_scores_per_class} indicate that the human perspective is not very different from the results achieved by state of the art computer vision systems~\cite{Chatfield-BMVC-14,Hinton-NIPS-2012}. Table~\ref{tab_scores_per_class} includes the mean Average Precision (mAP) performance of the best CNN classifier presented in~\cite{Chatfield-BMVC-14}. It can be observed that the lowest performance is obtained for the \emph{bottle}, \emph{potted plant} and \emph{chair} classes. These are also among the top $5$ most difficult classes for humans according to our findings. Moreover, \emph{aeroplane} and \emph{bird} are among the top $4$ easiest classes for both humans and machines.

\section{Learning to predict visual search difficulty}
\label{sec_prediction}
\vspace{-0.1cm}

So far, we obtained a set of ground-truth difficulty scores based on human annotations. We now go a step further and train a model to predict the difficulty of an input image.
We compare our supervised model with a handful of baseline models. 
We first describe our supervised model and the baseline models and then present experimental results.

\subsection{Our regression model}
\label{ssec_regression_model}
\vspace{-0.1cm}

We build our predictive model based on CNN features and linear regression with $\nu$-SVR~\cite{taylor-Cristianini-cup-2004} or Kernel Ridge Regression (KRR)~\cite{taylor-Cristianini-cup-2004}. We considered two pre-trained CNN architectures provided in~\cite{matconvnet}, namely VGG-f~\cite{Chatfield-BMVC-14} and VGG-verydeep-16~\cite{Simonyan-ICLR-14}. These CNN models are trained on the
ILSVRC benchmark~\cite{Russakovsky2015}.

We removed the last layer of the CNN models and used them to extract deep features as follows. The input image is divided into $1 \times 1$, $2 \times 2$ and $3 \times 3$ bins in order to obtain a pyramid representation for increased performance. The input image is also horizontally flipped and the same pyramid is applied over the flipped image. Finally, the $4096$ CNN features extracted from each bin are concatenated into a single feature vector for the original input image. The final feature vectors are normalized using the $L_2$-norm. The normalized feature vectors are then used to train either a $\nu$-SVR or a KRR model to regress to the ground-truth difficulty scores. We use our learned models as a continuous measure to automatically predict visual search difficulty.

\subsection{Baselines}
\label{ssec_baselines}
\vspace{-0.1cm}

We try out several baselines. Each baseline can assess the visual search difficulty based on some specific feature: image area, file size, objectness~\cite{objectness-PAMI-2012}, edge strengths~\cite{Dollar2015PAMI}, number of segments~\cite{Felzenszwalb2004IJCV}. Unlike the image properties analyzed in Section~\ref{sec_what_makes}, these features can be computed at test time (without manual annotations). 

\begin{figure}[t]
\begin{center}
\includegraphics[width=1.0\columnwidth]{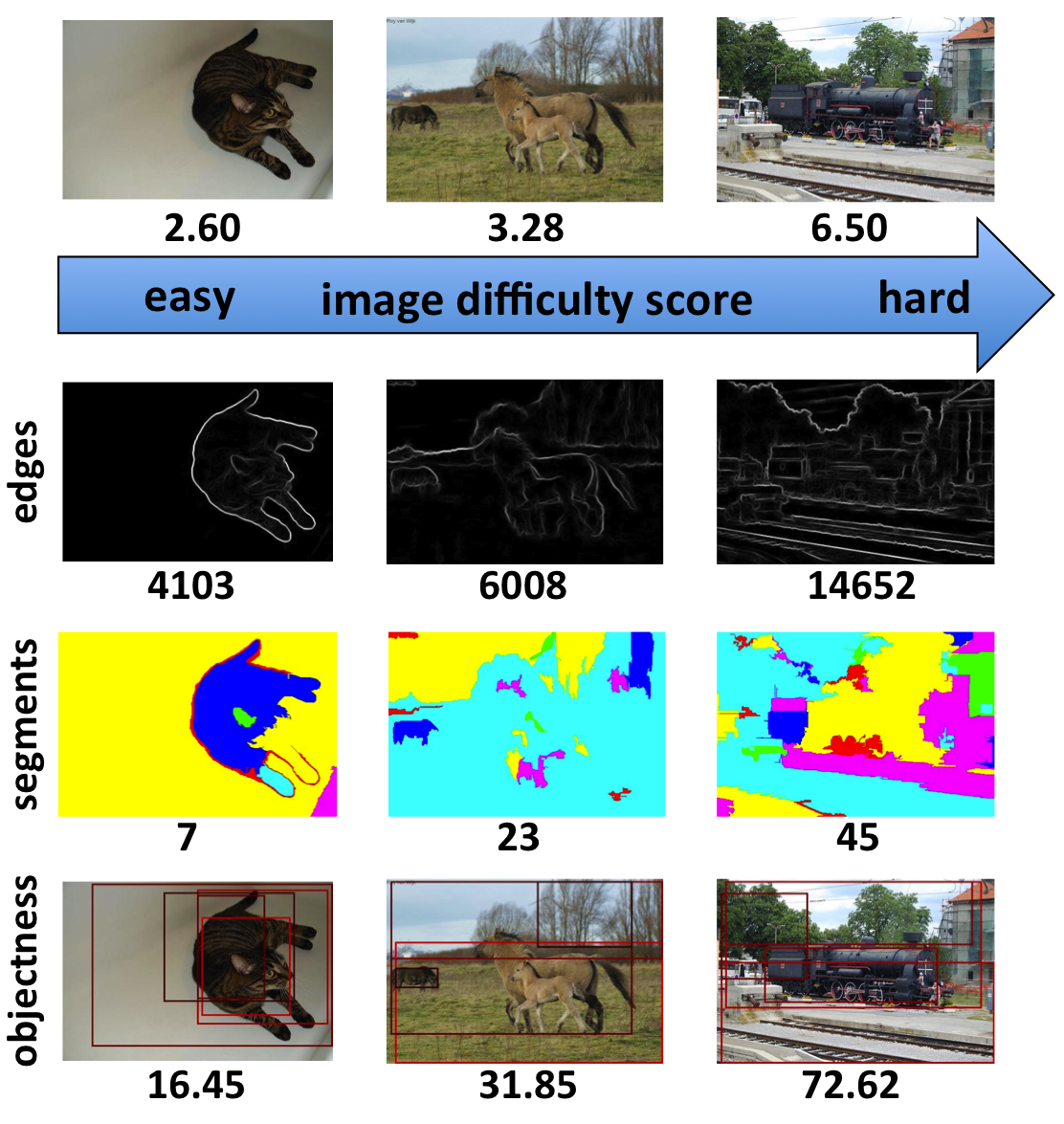}
\end{center}
\vspace*{-0.4cm}
\caption{Visual search difficulty assessed by baselines. We show the global image features used by each baseline for computing a difficulty score. Top row: input images with predicted scores by our method. Following rows: image edge maps~\cite{Dollar2015PAMI}, image segmentation~\cite{Felzenszwalb2004IJCV}, top $5$ highest scoring objectness windows (colored red to black from highest to lowest).}
\label{fig_features_scores}
\vspace*{-0.5cm}
\end{figure}

\noindent
{\bf Random scores.}
We assign random scores for each image.

\noindent
{\bf Image area.}
Without any prior information about the image content, the visual search task should be more difficult on larger images than on smaller ones. Based on this intuition, without analyzing the pixels inside an image, we quantify the difficulty of an image by its area.

\noindent
{\bf File size.}
A similar feature for quantifying visual search difficulty without looking at the pixels is the image file size. The images in PASCAL VOC 2012 are all compressed, in JPEG format. In this context, we tried recovering the compression rate induced to the original image by normalizing the file size with the image area, but it did not provide better results than the image file size alone.

\noindent
{\bf Objectness.}
The objectness measure~\cite{objectness-PAMI-2012} quantifies how likely it is for an image window to contain an object of any class. It is trained to distinguish windows containing an object with a well defined boundary and center, such as cows, cars and telephones, from windows covering amorphous background such as grass, road and sky. 
We used the official objectness code and obtained, for each image, a difficulty score by summing all the objectness scores of sampled windows. 
%
This difficulty score quantifies the image clutter through the objectness distribution in the $4D$ space of image windows. An easy image (Figure~\ref{fig_features_scores}, first column) should have a small score as it contains only a small number of windows with high objectness (red colored windows) covering the dominant object. All other windows, not covering objects, have small objectness (black colored windows). Conversely, a harder image (Figure~\ref{fig_features_scores}, last column) would have several peaks in the objectness distribution in the $4D$ space of image windows corresponding to objects' positions in the image.
We tried several variants for obtaining a difficulty measure by using objectness: (i) entropy of the objectness distribution estimated with kernel density in the $4D$ space of all image windows; (ii) mean value of the objectness heat map obtained by accumulating objectness scores at each pixel for all windows containing the respective pixel; (iii) entropy of the sampled objectness windows; (iv) sum of all (usually $1000$ samples obtained via the NMS sampling procedure~\cite{objectness-PAMI-2012}) objectness windows scores. We found out that all variants are essentially the same in terms of performance (Kendall's $\tau$ correlations between $0.20$ and $0.24$), with (iv) being marginally better.

\noindent
{\bf Edge strengths.}
Humans can easily find objects in cluttered scenes by detecting their contours~\cite{Shapley1973}. We use this idea to provide a measure of difficulty based on edges. Intuitively, an image with a smaller density of edges should be easier to search than another image with higher density. We use the fast edge detector of~\cite{Dollar2015PAMI} to compute the edge map of an image and characterize its visual search difficulty by the sum of edge strengths. 

\noindent
{\bf Segments.}
A different way of measuring difficulty rests on using segments as features. Segments divide an image into regions of uniform texture and color. Ideally, each segment should correspond to an object or to a background region. We quantify the complexity of an image by counting the number of segments. 
While turbo-pixels~\cite{Levinshtein2009PAMI} segment the image in regular small regions, essentially providing the same number of superpixels per image, the method of~\cite{Felzenszwalb2004IJCV} divides the image into irregular segments covering objects and larger portions of uniform background with fewer superpixels (Figure~\ref{fig_features_scores}). We use the available segmenter tool of~\cite{Felzenszwalb2004IJCV} with the default parameters for segmenting an image and characterize the difficulty by the number of segments.

\subsection{Experimental Analysis}
\label{ssec_experiments}
\vspace{-0.1cm}

\begin{table}[t]
\footnotesize{
\begin{center}
\begin{tabular}{|l|r|r|}
\hline
Model                                           & MSE               & Kendall $\tau$ \\
\hline
\hline
Random scores                                   & $0.458$           & $0.002$ \\
Image area                                      & -                 & $0.052$ \\
Image file size                                 & -                 & $0.106$ \\
Objectness~\cite{objectness-CVPR-2010,objectness-PAMI-2012}& -      & $0.238$ \\    
Edge strengths~\cite{Dollar2015PAMI}            & -                 & $0.240$ \\
Number of segments~\cite{Felzenszwalb2004IJCV}  & -                 & $0.271$ \\
\hline
Combination with $\nu$-SVR                      & $0.264$           & $0.299$ \\
\hline
VGG-f + KRR                                     & $0.259$           & $0.345$ \\
VGG-f + $\nu$-SVR                               & $0.236$           & $0.440$ \\
VGG-f + pyramid + $\nu$-SVR                     & $0.234$           & $0.458$ \\
VGG-f + pyramid + flip + $\nu$-SVR              & $0.233$           & $0.459$ \\
VGG-vd + $\nu$-SVR                              & $0.235$           & $0.442$ \\
VGG-vd + pyramid + $\nu$-SVR                    & $0.232$           & $0.467$ \\
VGG-vd + pyramid + flip + $\nu$-SVR             & $0.231$           & $0.468$ \\
VGG-f + VGG-vd + pyramid + flip + $\nu$-SVR     & $\mathbf{0.231}$           & $\mathbf{0.472}$ \\
\hline
\end{tabular}
\end{center}
}
\vspace*{-0.1cm}
\caption{Visual search difficulty prediction results of baseline models versus our regression models based on deep features extracted by VGG-f~\cite{Chatfield-BMVC-14} and VGG-verydeep-16 (VGG-vd)~\cite{Simonyan-ICLR-14}. KRR and $\nu$-SVR are alternatively used for training our model on $5,770$ samples from PASCAL VOC 2012. The mean squared error (MSE) and the Kendall's $\tau$ correlation are computed on a test set of $2,885$ samples. The best results are highlighted in bold.}
\label{tab_prediction_results}
\vspace*{-0.3cm}
\end{table}

\noindent
{\bf Evaluation measures.}
In order to evaluate the proposed regression model for predicting visual search difficulty, we report both the mean squared error (MSE) and the Kendall's $\tau$ rank correlation coefficient~\cite{upton-dict-stat-2008}. We report only the Kendall's $\tau$ correlation coefficient for the baseline models that do not involve regression, since the scores predicted by the baseline models are on a different range compared to the ground-truth difficulty scores and the MSE is a quantitative measure of performance unsuitable in this case.

\noindent
{\bf Evaluation protocol.}
We use the same split of the data set as described in Section~\ref{sec_what_makes}.
The validation set is used for tuning the regularization parameters of $\nu$-SVR and KRR. 

\noindent
{\bf Results.}
Table~\ref{tab_prediction_results} shows the results of different methods for predicting the ground-truth difficulty. Using random scores to assess difficulty leads to almost zero accuracy, showing that visual search difficulty estimation is not a trivial problem. Baselines that do not analyze image pixels perform a little bit better but are far away from accurately predicting the order of the images based on their difficulty. The methods based on mid-level features offer an increase in accuracy. Objectness and edge strengths perform essentially the same, achieving a correlation rank around $0.24$. Using segments further improves the performance to around $0.27$. Combining all these baselines with the $\nu$-SVR framework, we obtain a predictor that achieves a Kendall's $\tau$ rank correlation of about $0.30$. Based on the Kendall's $\tau$ definition, this translates in ranking about $65\%$ image pairs correctly. 

\begin{figure}[t]
\vspace*{-0.4cm}
\begin{center}
\includegraphics[width=0.95\columnwidth]{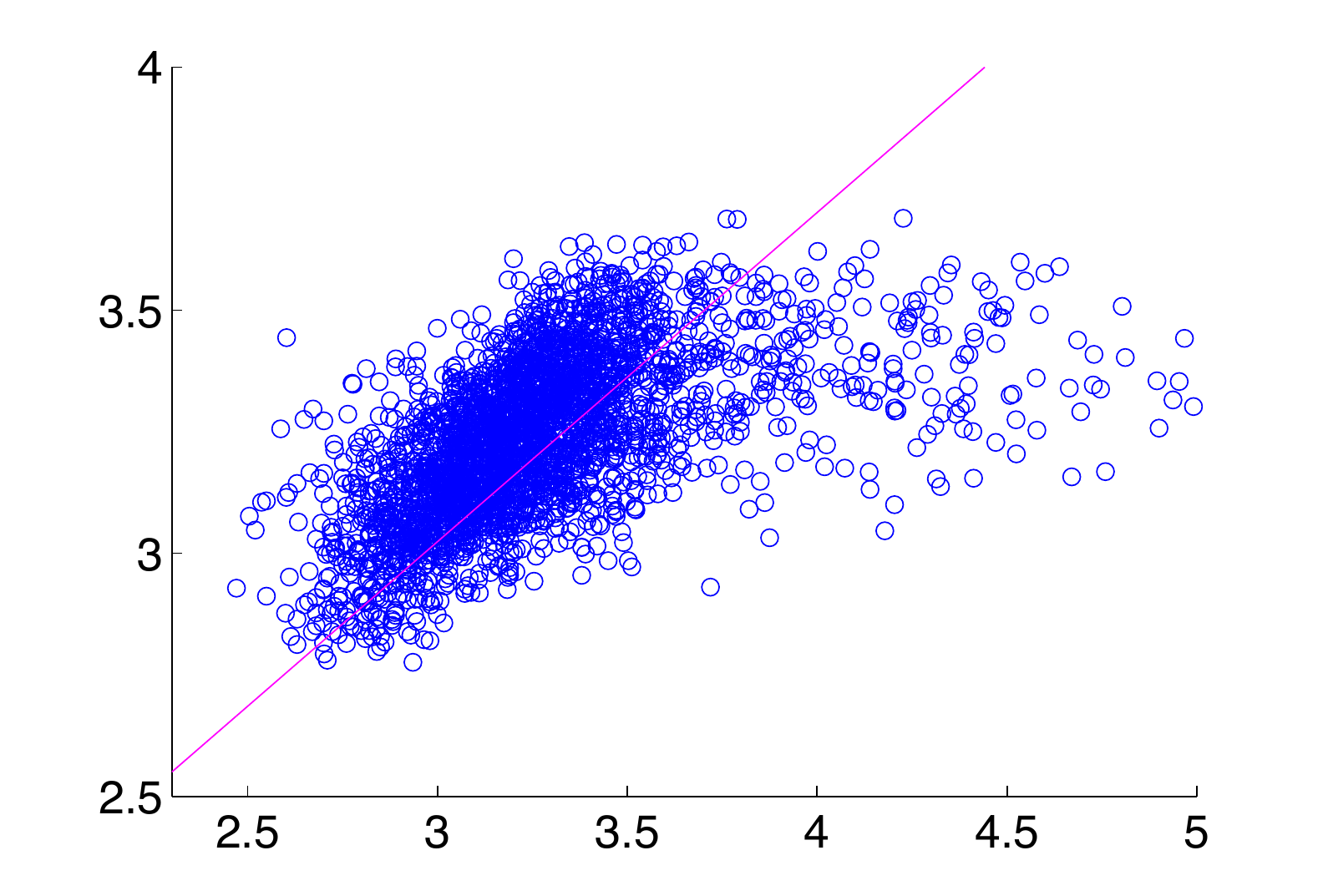}
\end{center}
\vspace*{-0.3cm}
\caption{Correlation between ground-truth (x-axis) and predicted (y-axis) difficulty scores. The least squares regression line is almost diagonal suggesting a strong correlation.}
\label{fig_predicted_vs_ground}
\vspace*{-0.3cm}
\end{figure}

\begin{table*}[t]
\footnotesize{
\begin{center}
\begin{tabular}{|l|c|c|c|c|c|c|c|c|c|}
\hline
Model        & Iteration 1 & Iteration 2 & Iteration 3 & Iteration 4 & Iteration 5 & Iteration 6 & Iteration 7 & Iteration 8 & Iteration 9             \\
\hline
\hline
Standard MIL  & 26.5\%    &	29.9\%    &	31.8\%    &	32.7\%    &	33.3\%    &	33.6\%    &	33.9\%    &	34.3\%  &	34.4\% \\
\hline
{\emph{Easy-to-Hard}} MIL & 31.1\%	&   36.1\%    &  	36.8\%    &	38.9\%    &	40.1\%    &	40.8\%    &	42.1\%    &	42.4\%  &	42.8\% \\
\hline
\end{tabular}
\end{center}
}
\vspace*{-0.1cm}
\caption{CorLoc results for standard MIL versus {\emph{Easy-to-Hard}} MIL.}
\label{tab_mil_results}
\vspace*{-0.4cm}
\end{table*}

In training our regression models, we tested out several configurations, including two neural network architectures (VGG-f and VGG-verydeep-16), various ways of extracting features (standard, pyramid, horizontal flip), and finally, two different regression methods, namely Kernel Ridge Regression and Support Vector Regression.
The least accurate configuration (VGG-f + KRR) gives already better performance compared to the baselines, reaching a rank correlation coefficient of $0.345$. Changing the regression method, $\nu$-SVR instead of KRR, we obtain a substantial increase to $0.440$. The best approach is to combine the pyramid features from both CNN architectures and to train the model using $\nu$-SVR. This combination outperforms by far all the baselines and their combination, and it remarkably achieves better performance than the image properties investigated in Section~\ref{sec_what_makes}, which require knowledge of the number objects, classes, bounding boxes (unavailable at test time). The best approach based on linear regression reaches a Kendall's $\tau$ correlation coefficient of $0.472$, which means that it correctly ranks about $75\%$ image pairs. 
We consider the best regression model as our difficulty predictor and use it in two applications in Section~\ref{sec_applications}.

Figure~\ref{fig_predicted_vs_ground} shows the correlation between the ground-truth and the predicted difficulty scores. The cloud of points forms a slanted Gaussian with the principal component oriented almost diagonally, indicating a strong correlation between the predicted and ground-truth scores. 
 
The examples presented in Figure~\ref{fig_easy_vs_medium_vs_hard} visually confirm the performance of our model: images with small number of objects and uniform backgrounds are ranked lower in difficulty than cluttered images with many objects and complex backgrounds. We explain the high accuracy of our model through the powerful features that capture visual abstractions at a higher level, close to the level of object class recognition. Since we define difficulty based on human response times for a visual search task that involves object detection and recognition, the fact that the best features are the higher level ones makes perfect sense. 
Analyzing the image content at lower levels (edge strengths, objectness, segmentation) is not good enough, showing that a higher level of interpretation is needed in order to assess difficulty.

\noindent
{\bf Machine versus human performance.} 
Interestingly, when our best difficulty predictor is evaluated
on the same $56$ images used for computing human agreement ($0.562$) in Section~\ref{ssec_measuring}, we obtain a Kendall's $\tau$ correlation of $0.434$.
Notably, our best difficulty predictor correctly ranks about $72\%$ image pairs, which is just a little lower than the average human performance of $80\%$ image pairs correctly ranked.

\noindent
{\bf Generalization across classes.} 
To demonstrate that our difficulty measure generalizes to classes not seen during training, we consider the setting where we train and test on disjoint PASCAL VOC 2012 classes. We train on 10 classes (\emph{bicycle}, \emph{bottle}, \emph{car}, \emph{chair}, \emph{dining table}, \emph{dog}, \emph{horse}, \emph{motorbike}, \emph{person}, \emph{TV monitor}) and test on the remaining 10 classes. We remove images containing both training and testing classes. The classes are split in order to exclude a minimal number of images ($1601$). 
In this setting, our $\nu$-SVR model based on CNN features obtains a Kendall's $\tau$ correlation of $0.427$, compared to $0.270$ for the $\nu$-SVR model that combines all the baselines.
This result is rather close to that obtained without separating classes ($0.472$).
Hence, this shows that our system \emph{generalizes well across classes}.

\section{Applications}
\label{sec_applications}
\vspace{-0.1cm}

We demonstrate the usefulness of our difficulty measure in two applications: weakly supervised object localization and semi-supervised object classification.

\subsection{Weakly supervised object localization}
\vspace{-0.1cm}

In a weakly supervised object localization (WSOL) scenario, we are given a set of images known to contain instances of a certain object class. In contrast to the standard full supervision, the location of the objects is unknown. The task is to localize the objects in the input images and to learn a model that can detect new class instances in a test image. 
%
Often, WSOL is addressed as a Multiple Instance Learning (MIL) problem~\cite{bilen14bmvc,cinbis14cvpr,deselaers-IJCV-2012,dietterich1997AI,shi12bmvc,siva11iccv,song14icml,song14nips}. In the MIL paradigm, images are treated as bags of windows (instances). A negative image contains only negative windows, while a positive image contains at least one positive window, mixed in with a majority of negative ones.
The goal is to find the true positives instances from which to learn a window classifier for the object class. This typically happens by iteratively alternating two steps: (i) select instances in the positive images based on the current window classifier; (ii) update the window classifier given the current selection of positive instances and all windows from negative images.

\noindent
{\bf Learning protocol.}
We employ our measure of difficulty as an additional cue in the standard MIL scheme for WSOL. We design a simple learning protocol that integrates the difficulty measure: rank input images by their estimated difficulty and pass them in this order to the standard MIL. We call this {\emph{Easy-to-Hard}} MIL.
%

\noindent
{\bf Evaluation protocol.}
We perform experiments on the training and validation sets of PASCAL VOC 2007 \cite{PASCAL07}. 
The main goal of WSOL is to localize the object instances in the training set. Following the standard evaluation protocol in the WSOL literature, we quantify this with the Correct Localization (CorLoc) measure 
\cite{cinbis14cvpr,deselaers10eccv,shi12bmvc,siva11iccv,wang15}. For a given target class, a WSOL method outputs one window in each positive training image. CorLoc is the percentage of images where the returned window correctly localizes an object of the target class according to the PASCAL VOC criterion (intersection-over-union $> 0.5$ \cite{PASCAL-voc-2012}). 

\noindent
{\bf Implementation details.}
We represent each image as a bag of windows extracted using the state-of-the-art
object proposal method of~\cite{dollar14eccv}. This produces about $2,000$ windows per image. Following~\cite{bilen14bmvc,girshick14cvpr,song14icml,song14nips,wang15}, we describe windows by the output of the second-last layer of the CNN model~\cite{Hinton-NIPS-2012}, pre-trained for whole-image classification on ILSVRC~\cite{Russakovsky2015}, using the Caffe implementation~\cite{jia2014caffe}. This results in $4096$-dimensional features. We employ linear SVM classifiers that we train with a hard-mining procedure at each iteration.
For our {\emph{Easy-to-Hard}} MIL we split the images in $k$ batches according to their difficulty. We use the easiest images (easiest batch) first, in order to update the window classifier, and progressively use more and more difficult batches. We used $k=3$ batches and $3$ iterations per batch, for a total of $9$ iterations.
The standard MIL baseline instead uses all images in every iteration.

\noindent
{\bf Results.}
In Table~\ref{tab_mil_results}, we compare the performance of our {\emph{Easy-to-Hard}} MIL with the standard MIL, in terms of average CorLoc over all $20$ classes. From the first iteration the improvement is already noticeable: almost $+5\%$ CorLoc. Easier images lead to a better initial class model as the MIL has a higher chance to detect class specific patterns and localize objects correctly.
The improvement increases as we add more batches: $+7\%$ after the second batch and $+8.4\%$ after the third. This increase demonstrates that the order in which images are processed is important in WSOL. Processing easier images in the initial stages results in better class models that in turn improve later stages.
Remarkably, our difficulty measure is trained on PASCAL VOC 2012, while here, we used it to quantify difficulty on images from PASCAL VOC 2007. As these two datasets have no images in common, the results show that our measure can \emph{generalize across different data sets}.
Finally, we point out that better CorLoc performance results have been reported on PASCAL VOC 2007 by other works using different WSOL algorithms~\cite{cinbis14cvpr,wang15}.

\subsection{Semi-supervised object classification}
\vspace{-0.1cm}
Here we use our difficulty measure in a second application, namely in predicting whether an image contains a certain object class (without localizing it).

\noindent
{\bf Learning protocol.}
We consider three sets of samples: a set $L$ of labeled training samples, a set $U$ of unlabeled training samples and a set $T$ of unlabeled test samples.
Our learning procedure operates iteratively, by training at each iteration a classifier on an enlarged training set $L$. We enlarge the training set at each iteration by moving $k$ samples from $U$ to $L$ as follows: we select $k$ samples from $U$ based on some heuristic, we label them (positive or negative) using the current classifier and move them from $U$ to $L$. We stop the learning when $L$ reaches a certain number of samples. The final trained classifier is tested on the test set $T$.

%


\noindent
{\bf Selection heuristics.}
To select the $k$ samples from $U$ at each iteration, we use one of the following heuristics:
(i) select the samples randomly (RAND);
(ii) select the easiest $k$ samples based on the ground-truth difficulty scores (GTdifficulty);
(iii) select the easiest $k$ samples based on the predicted difficulty scores (PRdifficulty);
(iv) select the most confident (farthest from the hyperplane) $k$ examples from $U$ according to the current classifier confidence score (HIconfidence);
(v) select the least confident (closest to the hyperplane) $k$ examples from $U$ according to the current classifier (LOconfidence);
(vi) select the least confident $K$ examples from $U$ according to the current classifier, and from these $K$, take the easiest $k$ examples based on our predicted difficulty score (LOconfidence+PRdifficulty).

\noindent
{\bf Evaluation protocol.}
We evaluate the classification performance of several models on PASCAL VOC 2012. All models are linear SVM classifiers based on CNN features~\cite{Simonyan-ICLR-14}.
We use as test set $T$ the official PASCAL {\emph{validation}} set, and we partition the PASCAL {\emph{train}} set into $L$ and $U$. We stopped the learning process when $L$ reached $3$ times more samples than the initial training set. We choose the initial $L$ to have $500$ labeled images randomly selected and repeat each run for $20$ times to reduce the amount of variation in the results. We report the mean Average Precision (mAP) performance. 
We set $k$ to $50$ and $K$ to $2000$.
In addition to the 6 models given by the above heuristics, we include a baseline model (\emph{BASIC}) trained only on the initial set $L$. We evaluate all models on the 7 classes (\emph{aeroplane}, \emph{bird}, \emph{car}, \emph{cat}, \emph{chair}, \emph{dog} and \emph{person}) from PASCAL VOC 2012 that include more than $5\%$ positive samples.
If the number of positive samples is not large enough, our semi-supervised learning protocol has trouble capturing feature patterns of the class. 


\begin{figure}[t]
\begin{center}
\includegraphics[width=0.96\columnwidth]{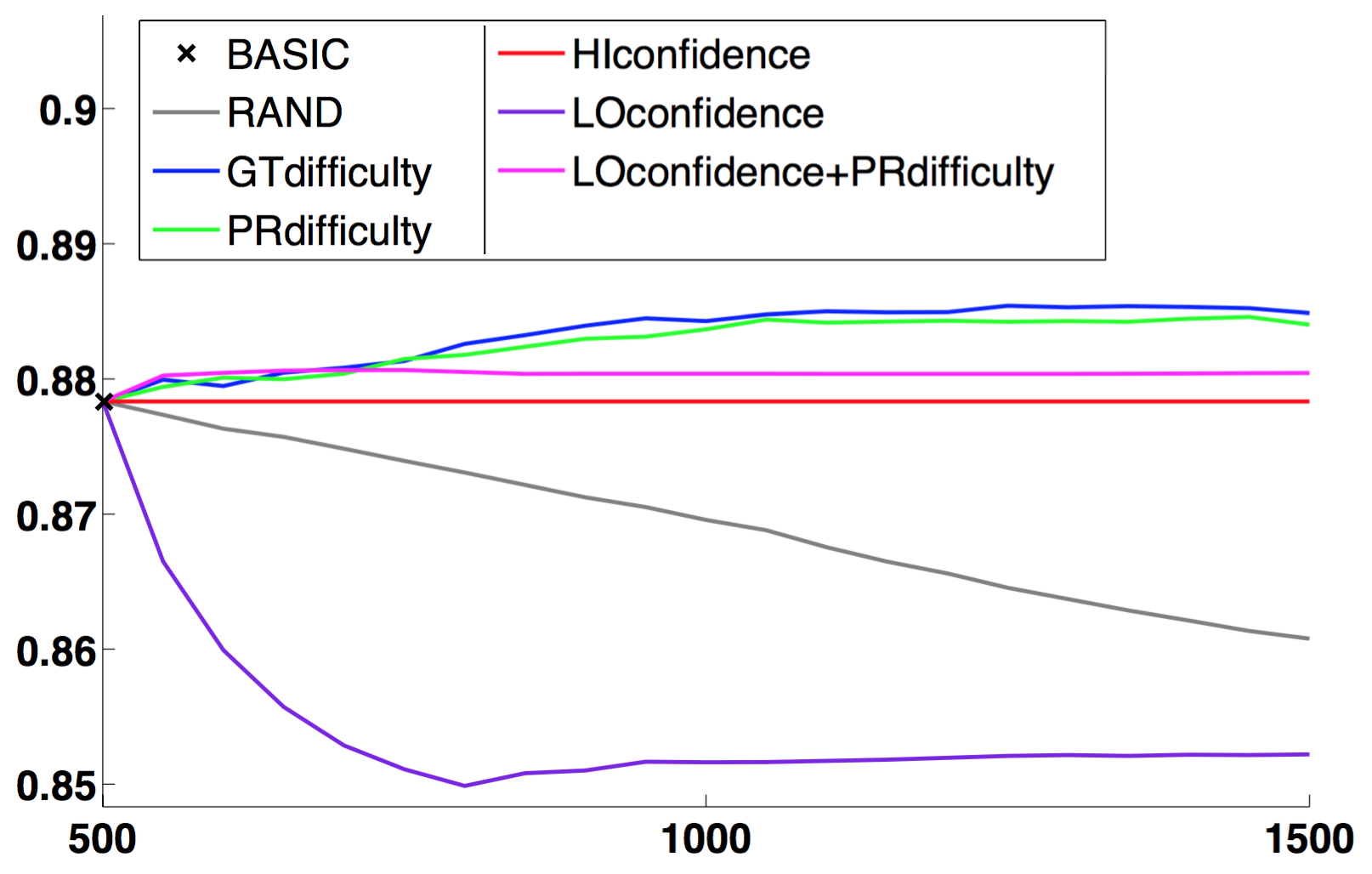}
\end{center}
\vspace*{-0.3cm}
\caption{The mAP performance (y-axis) as the size of the training set (x-axis) grows by adding automatically labeled samples using different heuristics (compared to the BASIC baseline).}
\label{fig_self_training}
\vspace*{-0.4cm}
\end{figure}

\noindent
{\bf Results.}
Figure~\ref{fig_self_training} shows the evolution of mAP for the proposed heuristics and the baseline.
Randomly choosing $50$ examples leads to a decrease in performance (86.1\% $\pm$ 1.0\%) compared to the BASIC method (87.8\% $\pm$ 0.6\%). Adding the most confident examples from $U$ (HIconfidence) does not influence the results  because the support vectors remain essentially the same. Using the least confident examples from $U$ (LOconfidence) in order to change the support vectors decreases performance (85.2\% $\pm$ 1.1\%).
The only useful information is provided by the difficulty scores, either predicted (88.4\% $\pm$ 0.6\%) or ground-truth (88.5\% $\pm$ 0.7\%), although it improves performance by less than $1\%$. Interestingly, by taking the least confident $2,000$ examples from $U$, and the easiest $50$ from these examples based on our predicted difficult score (LOconfidence+PRdifficulty), we can also improve performance (88.1\% $\pm$ 0.7\%) by a little margin. 

\vspace{-0.1cm}
\section{Future work}
\label{sec_conclusions}
\vspace{-0.1cm}

Curriculum learning~\cite{Bengio-ICML-2009} can help to optimize the training of deep learning models. We believe that our difficulty measure can be used in a curriculum learning setting to optimize the training of CNN models for various vision tasks.

\vspace{-0.3cm}
\subsubsection*{Acknowledgments}
\vspace{-0.1cm}
Vittorio Ferrari was supported by the ERC Starting Grant VisCul.
Radu Tudor Ionescu was supported by the POSDRU/159/1.5/S/137750 Grant.
Marius Leordeanu was supported by project number PNII PCE-2012-4-0581.

{\small
\bibliographystyle{ieee}
\bibliography{references}
}

\end{document}